\documentclass{article} 
\usepackage{nips11submit_e,times}
\usepackage{amsmath, amsthm}
\usepackage{amsfonts}
\usepackage{graphicx} 
\usepackage{subfigure} 
\usepackage{multirow} 
\usepackage{pifont} 
\usepackage[numbers]{natbib} 

\title{On Training Deep Boltzmann Machines}

\author{
Guillaume Desjardins,
Aaron Courville,
Yoshua Bengio \\
\texttt{\{desjagui,courvila,bengioy\}@iro.umontreal.ca} \\
D\'{e}partement d'informatique et de recherche op\'{e}rationnelle \\
Universit\'{e} de Montr\'{e}al \\
}

\nipsfinalcopy 

\begin{document}

\maketitle
\vspace*{-4mm}
\begin{abstract}
The deep Boltzmann machine (DBM) has been an important development in the quest
for powerful ``deep'' probabilistic models. To date, simultaneous or joint
training of all layers of the DBM has been largely unsuccessful with
existing training methods. We introduce a simple regularization scheme that
encourages the weight vectors associated with each hidden unit to have
similar norms. We demonstrate that this regularization can be easily combined with
standard stochastic maximum likelihood to yield an effective
training strategy for the simultaneous training of all layers of the deep
Boltzmann machine.
\end{abstract}

\section{Introduction}

Since its introduction by \citet{Salakhutdinov2009}, the deep Boltzmann
machine (DBM) has been one of the most ambitious attempts to build a probabilistic model with many layers of
latent or hidden variables. 
The DBM shares some characteristics with the earlier deep belief
network (DBN) \cite{Hinton06}. Both models may be viewed as multi-layer or ``deep''
extensions of the popular restricted Boltzmann machine (RBM). However,
unlike the DBN, commonly used DBM approximate inference schemes implement
true feedback mechanisms where the inferred activations of high-level units can
influence the activations of lower-level units. Thus alternative
interpretations of an input are able to compete at all levels of the model
simultaneously. This sort of sophistication in inference has the potential
to lead to more robust and globally coherent inferences, which are likely
to translate to better performance when the model is employed in tasks such
as classification.

Despite its success, the DBM has not yet entirely fulfilled this potential and the
DBN remains the more popular model paradigm for deep probabilistic models.
One potential explanation of why this is so could be tied to the difficulties one encounters when attempting
to estimate the model parameters from data.  Straightforward applications
of gradient-based methods such as stochastic maximum likelihood (SML)
\cite{Younes1999} (also known as persistent contrastive divergence
\cite{Tieleman08}) appear to fall in poor local minima, and fail to
adequately explore the space of model parameters.

One solution, as presented in \cite{Salakhutdinov2009}, is based on a greedy
layer-wise pretraining strategy which appears to overcome these poor local
minima . Each layer is trained as an RBM with the latent activations of the
layer below as its input.  These layers are then recombined by rescaling
the weights to account for the doubling of inputs into each intermediate
layer.  While this procedure seems to work well, it involves many steps and
is more complicated than would be ideal.  Also, more importantly, the
necessity of layer-wise pretraining makes it very difficult for the
organization of upper layers to influence the topological organization of
lower layers. For example, in cases where we would like to learn a DBM that
is not fully connected between each layer, it would potentially be very
desirable for the global connectivity pattern (eg. local receptive fields)
to influence the pattern of learned filters at all layers. Layer-wise
pretraining precludes this possibility.  If, on the other hand, one could
jointly train all layers of a deep Boltzmann machine simultaneously, then
the pattern of activation of the upper layer units would have an
opportunity to influence the weights trained at the lower layers via their
effect on lower-layer units activations.

In this paper, we describe a simple scheme for joint training of all layers of a
deep Boltzmann machine. Our strategy is based on the
observation that the poor local minima in which SML falls are characterized
by high variance in the norms of the weight vectors, particularly those
between the data and the first layer of hidden units. Our solution is to
simply add a regularization term to the standard maximum likelihood
training criterion that penalizes large differences in the norms of the
weight vectors, both within a layer and across neighbouring layers. Our
regularization is based on the intuition that successful models are those
where all units contribute roughly equally to the representation of the
data. This is a common principle that has previously been applied in the
form of sparsity penalties for RBM and DBN training
\cite{HonglakL2008}, where all hidden units are encouraged to be active over
an equal proportion of the data. Our application of this principal is somewhat
different, we do not explicitly control the activation of the hidden units,
instead we encourage equal influence of each \emph{active} hidden unit. 
We demonstrate, with experiments, both the failure of standard SML training
for DBMs and the effectiveness of our regularized-SML DBM training strategy.

\section{Boltzmann Machines}

A Boltzmann machine is defined as a network of
symmetrically-coupled binary stochastic units (random variables). These
stochastic units can be divided into two groups: (1) the \emph{visible}
units $v \in \{0,1\}^D$ that represent the data, and (2) the \emph{hidden}
units $h \in \{0,1\}^N$ that mediate dependencies between the visible units
through their mutual interactions. The pattern of interaction is specified
through the energy function:
\begin{equation}
E_{\mathrm{BM}}(v,h;\theta) = -\frac{1}{2}v^TUv-\frac{1}{2}h^TVh-v^TWh-b^Tv-d^Th,
\label{eq:boltzmann_energy}
\end{equation}
where $\theta = \{U,V,W,b,d\}$ are the model parameters which respectively
encode the visible-to-visible interactions, the hidden-to-hidden
interactions, the visible-to-hidden interactions, the visible
self-connections, and the hidden self-connections (also known as biases). To
avoid over-parametrization, the diagonals of $U$ and $V$ are set to zero.

The energy function
specifies the probability distribution over the joint space $[v,h]$, via the Boltzmann distribution:
\begin{equation}
P(v,h) = \frac{1}{Z(\theta)}\exp\left(-E_{\mathrm{BM}}(v,h;\theta)\right),
\end{equation}
where the partition function $Z(\theta)$ ensures that the density
normalizes:
\begin{equation}
Z(\theta) = \sum_{v_1 = 0}^{v_1 = 1}\cdots \sum_{v_D = 0}^{v_D = 1}
\sum_{h_1 = 0}^{h_1 = 1}\cdots \sum_{h_N = 0}^{h_N = 1} \exp\left(-E_{\mathrm{BM}}(v,h;\theta)\right).
\end{equation}
This joint probability distribution gives rise to the set of conditional
distributions of the form:
\begin{eqnarray}
P(h_i \mid v, h_{\setminus i}) & = &
    \mathrm{sigmoid}
    \left( 
        \sum_{j}W_{ji}v_{j} + \sum_{i' \setminus i} V_{ii'} h_{i'} + d_{i} 
    \right) \\ 
P(v_j \mid h, v_{\setminus j}) & = &
    \mathrm{sigmoid}
    \left( 
        \sum_{i}W_{ji}v_{j} + \sum_{j' \setminus j} U_{jj'} v_{j'} + b_{j} 
    \right).
\end{eqnarray}
In general, inference in the Boltzmann machine is intractable. For example,
computing the conditional probability of $h_i$ given the visibles, $P(h_i
\mid v)$, requires marginalizing over the rest of the hiddens which implies
evaluating a sum with $2^{N-1}$ terms:
\begin{equation}
P(h_i \mid v) = \sum_{h_1 = 0}^{h_1 = 1}\cdots \sum_{h_{i-1} = 0}^{h_{i-1} = 1}
\sum_{h_{i+1} = 0}^{h_{i+1} = 1} \cdots \sum_{h_N = 0}^{h_N = 1} P(h \mid v)
\end{equation}
However with some judicious choices
in the pattern of interactions between the visible and hidden units, more
tractable subsets of the model family are possible.
 
\subsection{Restricted Boltzmann Machines}

The restricted Boltzmann machine (RBM) is likely the most popular subset of
Boltzmann machines. They are defined by restricting the interactions in the
Boltzmann energy function, in Eq. \ref{eq:boltzmann_energy}, to only those
between $h$ and $v$, i.e. for $E_{\mathrm{RBM}}$, $U=\mathbf{0}$ and $V=\mathbf{0}$. As such, the
RBM can be said to form a \emph{bipartite} graph with the visibles and the
hiddens forming two layers of vertices in the graph. With this
restriction, the RBM possesses the useful property that the conditional
distribution over the hidden units factorizes given the visibles:
\begin{equation}
P(h \mid v) = \prod_{i} P(h_{i} \mid v) = \prod_{i} \mathrm{sigmoid}\left(
  \sum_{j} W_{ji} v_{j}+d_{i}\right)
\end{equation}
Likewise, the conditional distribution over the visible units given the
hiddens also factorizes:
\begin{equation}
P(v \mid h) = \prod_{j} P(v_{j} \mid h) = \prod_{j} \mathrm{sigmoid}\left(
  \sum_{i} W_{ji} h_{i}+b_{j}\right)
\end{equation}
This conditional factorization property of the RBM immediately implies that
most inferences we would like make are readily tractable. For example, the
conditional independence of the hiddens implies that posterior marginals of the form
$P(h_{i} \mid v)$ are immediately available.

Importantly, the tractability of the RBM does not extend to its partition
function, which still involves sums with exponential number of terms.
It does imply however that we can limit the number of terms to
$\min\{2^D,2^N\}$. Usually this is still an unmanageable number of terms
and therefore we must resort to approximate methods to deal with its
computation.

Learning in the RBM is also rendered much more tractable in comparison to
the general Boltzmann machine. Typically, learning involves finding a set
of model parameters that approximately  maximizes the log likelihood of a 
training dataset:
\begin{equation}
\sum_{t} log P(v_{:,t}; \theta) = \sum_{t} log \sum_{h_{1,t}=0}^{h_{1,t}=1}\cdots\sum_{h_{N,t}=0}^{h_{N,t}=1}P(v_{:,t}, h_{:,t} ; \theta).
\end{equation}
This can be accomplished via gradient ascent. The gradient of the log likelihood
of the data for the RBM is given by:
\begin{equation}
\frac{\partial}{\partial\theta_{i}}\left(\sum_{t=1}^{T}\log p(v_{:,t})\right)
 = - \sum_{t=1}^{T}\left\langle
     \frac{\partial}{\partial\theta_{i}}E_{\mathrm{RBM}}(v_{:,t},h)\right\rangle _{p(h_{:,t}\mid v_{:,t})} + 
     \sum_{t=1}^{T}\left\langle
     \frac{\partial}{\partial\theta_{i}}E_{\mathrm{RBM}}(v,h)\right\rangle
   _{p(v,h)}, \nonumber
\end{equation}
where we have the expectations with respect to $p(h_{:,t}\mid v_{:,t})$
in the {}``clamped'' condition, and over the full joint $p(v_{:,t},h_{:,t})$
in the {}``unclamped'' condition.
In training, we follow the stochastic maximum likelihood (SML) algorithm (also know as
persistent contrastive divergence or PCD)~\cite{Younes1999,Tieleman08},
i.e., performing only one or a few updates of an MCMC chain between
each parameter update.

\section{Deep Boltzmann Machines}

\begin{figure}
   \centering
   \includegraphics[scale=0.50]{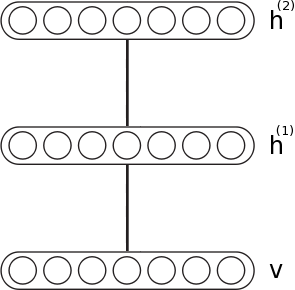}
   \caption[]{\small{\label{fig:dbm} A graphical depiction of a 2-layer DBM.
   Undirected weights connect the visible layer to layer $h^{(1)}$; and layers
   $h^{(1)}$ to $h^{(2)}$. Note that $v$ and $h^{(2)}$ are conditionally
   independent given $h^{(1)}$.}}
\end{figure}

The Deep Boltzmann Machine (DBM) is another
particular subset of the Boltzmann machine family of models where the units
are again arranged in layers. However unlike the RBM, the DBM possesses multiple
layers of hidden units as illustrated in Figure \ref{fig:dbm}. With
respect to the Boltzmann energy function of Eq.~\ref{eq:boltzmann_energy}, the DBM
corresponds to setting $U=0$ and a sparse connectivity structure in both
$V$ and $W$. We can make the structure of the DBM more explicit by specifying
its energy function. For the 2-layer model it is given as:
\begin{equation}
E_{\mathrm{DBM}}(v, h^{(1)}, h^{(2)}; \theta) = 
    -v^T W h^{(1)} - 
    \left.h^{(1)}\right.^T V h^{(2)} -
    \left.d^{(1)}\right.^T h^{(1)} -
    \left.d^{(2)}\right.^T h^{(2)} -
    b^T v ,
\label{eq:boltzmann_energy}
\end{equation}
with $\theta = \{W,V,d^{(1)},d^{(1)},b\}$. The DBM can also be characterized as
a bipartite graph between two sets of vertices, formed by the units in odd and even-numbered
layers (with $v:=h^{(0)}$).

\subsection{Mean-field approximate inference} 
A key point of departure from the RBM is that 
the posterior distribution over the hidden units (given the visibles) is no longer
tractable, due to the interactions between the hidden units. \citet{Salakhutdinov2009} resort to a
mean-field approximation to the posterior. Specifically, in the case of the
2-layer model, we wish to approximate $P\left(h^{(1)},h^{(2)} \mid v\right)$ with the factored
distribution $Q_v(h^{(1)},h^{(2)}) =
\prod_{j=1}^{N_1}Q_v\left(h^{(1)}_{j}\right)\
\prod_{i=1}^{N_2}Q_v\left(h^{(2)}_{i}\right)$. such that the KL
divergence $\mathrm{KL} \left( P\left(h^{(1)},h^{(2)} \mid v\right) \Vert
Q_v(h^{1},h^{2}) \right)$ is minimized or equivalently, that a lower
bound to the log likelihood is maximized:
\begin{equation}
\log P(v) > \mathcal{L}(Q_{v}) \equiv
\sum_{h^{(1)}}\sum_{h^{(2)}}Q_v(h^{(1)},h^{(2)})\log\left(\frac{P(v,h^{(1)},h^{(2)})}{Q_v(h^{(1)},h^{(2)})}\right)
\label{eq:variational_lower_bound}
\end{equation}

Maximizing this lower-bound with respect to the mean-field distribution
$Q_v(h^{1},h^{2})$ yields the following mean field update equations:
\begin{align}
    \label{eq:mf1}
    \hat{h}^{(1)}_i &\leftarrow \mathrm{sigmoid}\left( 
        \sum_j W_{ji} v_j + \sum_k V_{ik} \hat{h}^{(2)}_k + d^{(1)}_i
        \right) \\
    \label{eq:mf2}
    \hat{h}^{(2)}_k &\leftarrow \mathrm{sigmoid}\left( 
        \sum_i V_{ik} \hat{h}^{(1)}_i + d^{(2)}_k 
        \right)
\end{align}

Iterating Eq.~(\ref{eq:mf1}-\ref{eq:mf2}) until convergence yields the
parameters used to estimate the ``variational positive phase'' of
Eq.~\ref{eq:variational_gradient}:
\begin{align}
\mathcal{L}(Q_{v}) &=
    \mathbb{E}_{Q_v}\left[ 
    \log P(v,h^{(1)},h^{(2)}) -
    \log Q_v(h^{(1)},h^{(2)}) \right] \nonumber \\
&= \mathbb{E}_{Q_v} \left[ 
    - E_{\mathrm{DBM}}(v,h^{(1)},h^{(2)}) 
    - \log Q_v(h^{(1)},h^{(2)}) \right]
    - \log Z(\theta) \nonumber \\
\label{eq:variational_gradient}
\frac{\partial \mathcal{L}(Q_{v})} {\partial \theta} &= 
    -\mathbb{E}_{Q_v} 
    \left[ 
    \frac{\partial E_{\mathrm{DBM}}(v,h^{(1)},h^{(2)})}{\partial \theta} 
    \right]
    + \mathbb{E}_{P}
    \left[
    \frac{\partial E_{\mathrm{DBM}}(v,h^{(1)},h^{(2)})}{\partial \theta}
    \right]
\end{align}

Note that this variational learning procedure leaves the ``negative phase''
untouched. It can thus be estimated through SML or Contrastive Divergence
\cite{Hinton-PoE-2000} as in the RBM case.

\subsection{Training Deep Boltzmann Machines}
Despite the intractability of inference in the DBM, its training should not, in
theory, be much more complicated than that of the RBM. The major difference
being that instead of maximizing the likelihood directly, we instead choose
parameters to maximize the lower-bound on the likelihood given in
Eq.~\ref{eq:variational_lower_bound}. The SML-based algorithm for maximizing
this lower-bound is as follows:
\begin{enumerate}
\item Clamp the visible units to a training example.
\item Iterate over Eq.~(\ref{eq:mf1}-\ref{eq:mf2}) until convergence.
\item Generate negative phase samples $v^-$, $h^{(1)-}$ and $h^{(2)-}$ through SML.
\item Compute $\partial \mathcal{L}(Q_v) \left/ \partial \theta \right.$
    using the values obtained in steps 2-3.
\item Finally, update the model parameters with a step of gradient ascent.
\end{enumerate}

While the above procedure appears to be a simple extention of the highly
effective SML scheme for training RBMs, as we demonstrate in
Sec.~\ref{sec:exp}, this procedure seems vulnerable to falling in poor
local minima which leave many filters effectively dead (not significantly
different from its random initialization with small norm). 
\label{sec:layerwise}

The failure of the SML joint training strategy was noted by
\citet{Salakhutdinov2009}. As a far more successful alternative, they
proposed a greedy layer-wise training strategy.
This procedure consists in pre-training
the layers of the DBM, in much the same way as the Deep Belief Network: i.e.
by stacking RBMs and training each layer to independently model the output of
the previous layer.
A final joint ``fine-tuning'' is done following the above SML-based procedure.



\section{Joint Training of a DBM}
\label{sec:joint_training}

While greedy layer-wise training has been shown to be reasonably
successful, a means of joint training a DBM would be highly desireable. Not
only would it be simpler, but it would open the door to local
receptive field learning and more general architectures where we would
like the top-down pattern of connectivity to influence the learning of
lower-level features. In this section we detail our simple proposal for a means
to jointly train all layers of a DBM.

Our strategy is based on the observation that standard SML-based joint
training tends produce high variance in the weight vectors associated with
each hidden unit -- particularly across the first-layer weight vectors that
connect the visible units to the first hidden layer units. 

Our proposal is thus to regularize the maximum likelihood objective, in order
to encourage units to have similar norms (i) within a given layer and (ii) to a
lesser extent, across neighbouring layers. We thus introduce an additional parameter
$\mu^{(l)}$ for each layer of the DBM, representing the average norm of weight
vectors of the $l$-th layer. We then add a regularization term to our objective
$\mathcal{L}(Q_v)$, which penalizes deviations from this mean-value using a
squared-error penalty.  $\mu^{(l)}$'s belonging to adjacent layers are further
constrained to be close (again in the $l2$ sense). This gives rise to the
following regularized objective:
\begin{align}
    \label{eq:regll}
    \max_\theta \mathcal{L}(Q_{v}) +  \
    \alpha^{(1)} \sum_{j=1}^{N_1} \left( \Vert W_{\cdot i} \Vert_2 - \mu^{(1)} \right)^2 +\
    \alpha^{(2)} \sum_{i=1}^{N_2} \left( \Vert V_{\cdot j} \Vert_2 - \mu^{(2)} \right)^2 +\
    \gamma \left( \mu^{(1)} - \mu^{(2)} \right)^2.
\end{align}

One can think of this regularization term as a spring, which ensures that the
system evolves jointly from an initial random weight configuration (where the
columns of $W^{(l)}$ have small norm) to a good model of the input distribution
$P(v)$, which undoubtedly requires weight vectors with larger norms.

\section{Experiments}
\label{sec:exp}

To evaluate the effect of our regularization term, we trained deep Boltzmann
machines on the pervasive MNIST dataset \cite{LeCun98-small}. All our models
were trained for $10^6$ updates, using minibatches of size $50$, varying the
learning rate in $\{10^{-2},10^{-3},10^{-4}\}$.

\begin{figure}
    \centering
    \subfigure[Layer 1]
    {
        \label{fig:bad_l1}
        \includegraphics[scale=0.5]{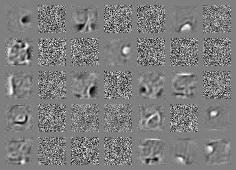}
    }
    \subfigure[Layer 2]
    {
        \label{fig:bad_l2}
        \includegraphics[scale=0.5]{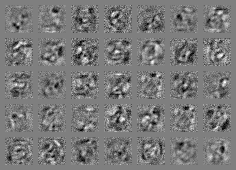}
    }
    \subfigure[Layer 3]
    {
        \label{fig:bad_l3}
        \includegraphics[scale=0.5]{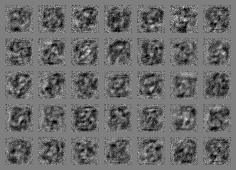}
    }
    \caption[]{\small{Random subset of filters obtained by a 784-500-500-1000
    DBM, after $10^6$ updates (trained jointly, using the direct approach). The
    learning rate was set to $10^{-3}$ and the batch size to $50$. Higher-level weights
    are visualized by performing a linear combination of lower-level weights,
    but only picking the $20$ most active connections.
    \label{fig:bad_filters}}}
\end{figure}

\paragraph{Direct Approach} 
Our first model was a 3-layer DBM with [500,500,1000] hidden units in the
first, second and third layers (respectively).  The resulting filters are shown
in Figure~\ref{fig:bad_filters}.

As we can see, many of the first-layer
filters are not significantly different from their random initialization, with
very small norm. Based on the difference between these results and the
successful training of an RBM, we speculate that the top-down interactions
prevent the first layer from learning useful filters. We have confirmed that
the high-frequency ``noise'' filters are actually the ones with lowest norm. It
seems as though early top-down input is influencing the activations of these
hidden units, perhaps directly through some form of suppression, or indirectly,
by reinforcing the activation of the subset of filters which become useful early on.
While we plot filters for 3-layer networks, the same results hold for networks
of depth two.


\paragraph{Joint Training}

\begin{figure}
    \centering
    \subfigure[Layer 1]
    {
        \label{fig:good_l1}
        \includegraphics[scale=0.5]{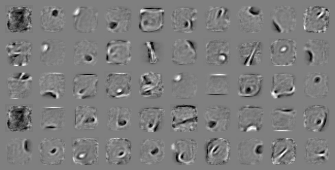}
    }
    \subfigure[Layer 2]
    {
        \label{fig:good_l2}
        \includegraphics[scale=0.5]{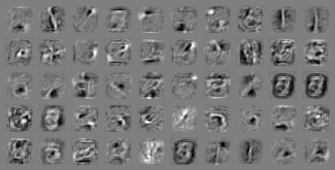}
    }
    \caption[]{\small{Random subset of filters obtained by a 784-500-500 DBM,
    after $10^6$ updates. All layers were trained jointly using the
    regularized cost of Eq.\ref{eq:regll}. The learning rate was set to
    $10^{-2}$ and the batch size to $50$.
    \label{fig:good_filters}}}
\end{figure}

Using our regularized objective, we train a 2-layer DBM with 500 hidden
units in each layer. We set the hyper-parameters of Eq.\ref{eq:regll} as
follows: $\alpha^{(1)}=0.1$, $\alpha^{(2)}=0.1$ and $\gamma = 1$. Furthermore,
we dampen the learning rate on the parameters $\mu^{(1)}$ and $\mu^{(2)}$ by a
factor of a thousand, in order for the layer norms to evolve more slowly.
A random subset of filters is shown in Figure~\ref{fig:good_filters}.

The effect of our regularization term is drastic: the vast majority of first
layer filters seem to train successfully and resemble the pen-stroke detectors
characteristic of features learned on MNIST.  While more difficult to
interpret, the second layer filters are also clearly superior to the ones of
Figure~\ref{fig:bad_filters}, combining lower-level features (pen strokes) into
more global digit-like objects.





\section{Discussion}

We have introduced a simple regularization scheme that appears to prevent
SML from falling into a poor local minimum of our objective function. The
regularizer encourages units to learn filters which have similar norms, both
within and across adjacent layers.
We have empirically demonstrated the failure of joint training of all
layers of a deep Boltzmann machine which leaves many filters in the lower
layer of a 2-layer DBM with very small norm and, consequently, little
contribution to modeling the data. We have also empirically demonstrated
the success of our regularization scheme in simultaneously training both layers of a
2-layer DBM. While we have not shown it here, our scheme also appears to
work well for DBMs with more than 2 layers. In future work, we would like
to determine how many layers can be learnt simultaneously using similar basic
norm-regularizaiton schemes.

\subsubsection*{Acknowledgments}

The authors would like to acknowledge the support of the following agencies for
research funding and computing support: NSERC, Calcul Qu\'{e}bec and CIFAR. We
would also like to thank the developers of Theano \cite{bergstra+al:2010-scipy}.

\small{
\bibliography{strings,strings-shorter,ml,aigaion-shorter}
\bibliographystyle{natbib}
}

\end{document}